# Review on Social Behavior Analysis of Laboratory Animals: From Methodologies to Applications

**Ziping Jiang[1], Paul L. Chazot[2] and Richard Jiang[1]**

[1] School of Computing and Communications, Lancaster University, InfoLab21, South Dr, Lancaster University, Bailrigg, Lancaster LA1 4WA, United Kingdom

[2] Department of Biosciences, Durham University, Durham, UK

## 1. Introduction

Behavior describes the way animals interact with each other as well as the environment. The study of animal behavior is a long-standing topic that can provide insights into various objectives. In phylogenetics, the behavior pattern is a critical feature for comparing and grouping different species, especially laboratory animals and microorganisms. In ethology, animal behaviors are recorded to distinguish fixed action patterns and evolve behaviors to investigate the effects of environments on organisms. Additionally, behavior analysis is also essential in genetic research. With the progress of biotechnology, state-of-the-art genetic engineering technologies can transfer genes within and across species boundaries to produce improved organisms.

The behavioral differences between experimental groups and comparison group are a determining factor for analyzing the underlying genotype of a modified organism. The most commonly used laboratory animals are fruit fly and house mice, also known as *Drosophila melanogaster* and *Mus musculus*. Fruit fly can exhibit a wide range of complex social behaviors through only neurons, making them an ideal target for test genetic research. House mouse, on the other hand, has a similar genotype to that of humans; therefore, it is widely used for research in psychology, medicine, and other scientific disciplines. However, the labeling behavior of laboratory animals is a labor-intensive job. First, as the behaviors are performed randomly, most datasets contain dozens of hours of videos to provide a statistically solid result. Second, most of the videos are required to be annotated frame-by-frame due to the fast speed and short duration of laboratory animal behaviors. At last, this job requires professional knowledge that only biologists with related work experience can fulfill. To alleviate their burden, computer vision approaches are then introduced for automatic annotating.

Automating the analysis of animal behaviors, similar to that of human behavior, includes tracking and detecting. At different stages of the progress of artificial intelligence research, the proposed approaches are based on different techniques. In this work, we aim to provide a thorough investigation of related work, furnishing biologists with a scratch of efficient animal behavior annotation methods. Apart from that, we also discuss the strengths and weaknesses of those algorithms to provide some insights for those who are engaged in this field.

This review is organized as follows. In Section 2, we list the remarkable dataset along with the original objective when proposed. From Sections 3 to 5, we provide a general introduction on methodologies of related topics, including feature extraction, statistical learning, and deep learning. In Sections 6 and 7, we review the application of these methods in tracking and detecting the behavior of laboratory animals. At last, Section 8 talks about the miscellaneous works that are relevant to the topic.

## 2. Datasets

The Caltech Resident-Intruder Mouse dataset (CRIM13) [4] is introduced for the study of mice social behavior involved in aggression and courtship. The dataset consists of 237 videos, recorded from side view and top, synchronously. The scene of each video is similar. At the beginning of each video, a male resident mouse stays in the cage, and at some point, an intruder is introduced into the experiment to investigate the social interaction between the mice. The actions contained in the video are categorized into 12 specified actions and one category to label the frames where no behavior of interest occurs. Mazur-Milecka and Ruminski [33] introduced a small dataset using for analysis of rat aggressive behavior. Instead of using the visual camera, they recorded the rat movement with a thermal camera. Since the cage is made from plexiglass without nesting, water nor food placed within, the heat of rats can be observed clearly.

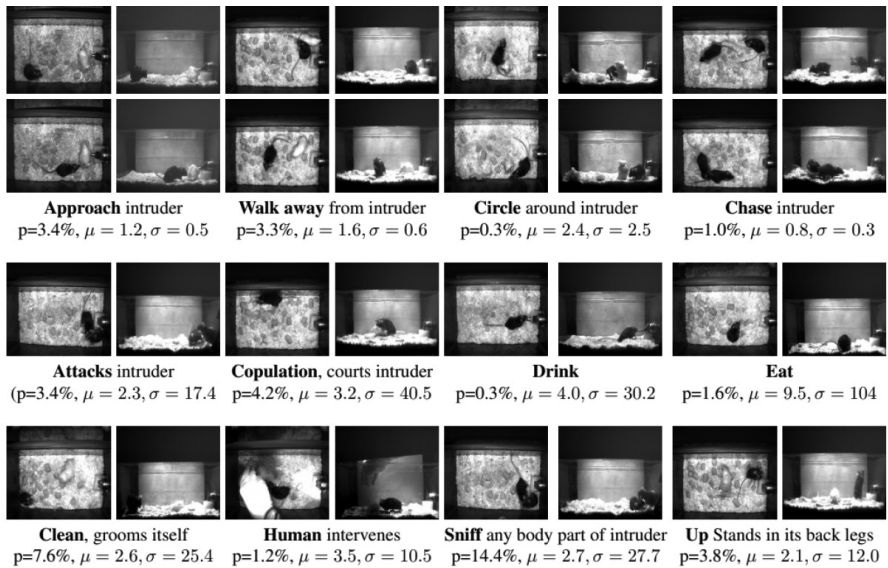

**Fig. 1:** Sample image from CRM13. Behavior categories: frame examples, descriptions, frequency of occurrence, ($p$), and duration mean and variance expressed in seconds ($\mu$, $\sigma$). All behaviors refer to the cage resident, main mousey need to use the optional argument [t] with the side caption command.

The objectives of most proposed fruit fly datasets are to identify the phenotype differences between genetically edited objects. The Rubin GAL4 collection [20] is one of the largest datasets that provide a variety of phenotypes caused by different transgenic lines. In particular, the dataset has 14,524 16-minute videos of 2,144 different lines. However, the minor behavior differences can hardly be detected by current techniques, therefore the Rbuin GAL4 collection is only adopted in tracking tasks. On the contrary, Caltech fly-vs-fly interactions [11] are proposed for detecting aggressive behaviors of genetically edited flies. It contains 22 hours of fruit fly social behaviors, including courtship, threat, tussle, etc. The dataset is split into three subsets, according to the gender of the test subjects. Along with the dataset, the authors also provide a feature detector system that is able to extract the feature representations. FlyBowl [40] records videos of groups of flies freely behaving in an open-field walking arena. The system is designed with automated tracking and behavior analysis algorithms for observing the social behaviors of multiple flies in a complex dynamic environment.

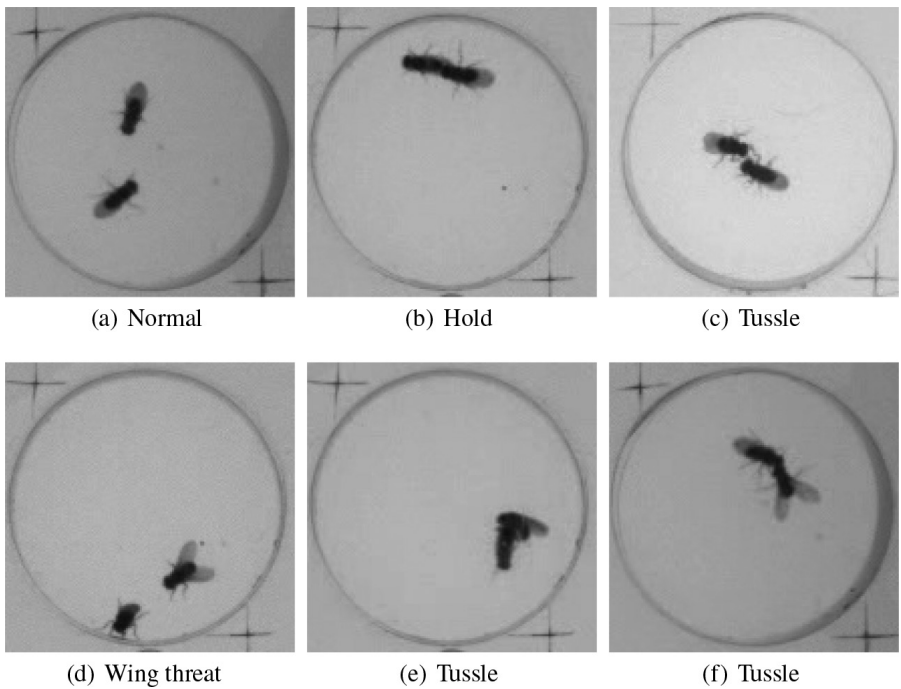

|  (a) Normal | (b) Hold | (c) Tussle |

|  (d) Wing threat | (e) Tussle | (f) Tussle |

**Fig. 2:** Examples of fly behaviors from fly-vs-fly [11]. (a) Normal flies without social behavior; (b) hold behavior between flies; (c) tussle behavior between flies; (d) wing threat behavior of a fly; (e) confusable tussle behavior (against hold); (f) confusable tussle behavior (against wing threat)

## 3. Methodologies

Research in tracking and labeling laboratory animal behavior has a long history. As an empirical topic, the methodologies of proposed models vary at different periods. In this section, we provide a brief discussion of related works, including (1) computer vision-based feature extraction methods that gathering useful information from raw data, (2) statistical learning methods that classify actions according to obtained feature, and (3) deep learning methods that provide an end-to-end solution to the tasks using a deep model with high computational power.

### 3.1 Feature Extraction

Intuitively, video data record the information about the changes of objects as well as the environment. To provide a convincing prediction of the recorded action in a sequence of frames, spatial features as well as temporal features should be taken into account.

Optical flow is one of the earliest approaches [8, 1]. It measures the apparent motion of objects in a sequence of frames. However, the validity of such measurement depends on the recording conditions. In fact, it can be influenced by variations in the background as well as the burst movements of multiple objects, especially when they are in occlusion.

Spatio-temporal interest point detection [26, 46]; on the other hand, focus on the points with high neighborhood variations in video, regardless of the size, frequency, and velocity of moving patterns. For those some of the data where useful information is also contained in the background, a descriptor is introduced for illustrating the surrounding of the interest points [27, 25].

### 3.2 Statistic Learning and Tracking Algorithms

Statistic learning dominates the research of Artificial Intelligence for decades. One of the most appealing properties of statistical learning methods is their explain ability and transparency. In particular, given a trained model $\mathcal{H}(\theta)$ and a feature vector $x = (x_1, x_2, \ldots, x_n)$, one can find out how the predicted result $\mathcal{H}(x, \theta)$ is influenced by each of the features. Moreover, since the training and prediction process can be observed, the model is open for amendment once the result is not satisfying.

The limitations of statistical learning methods are also straightforward. First, their capability is limited, since most of the real-world problems are too complex to be described by a model. Second, to obtain a satisfying result, the inputs of a model should be cautiously designed, indicating the pre-processing of data that requires great effort. At last, as a natural consequence of the high requirement of pre-processing, it lacks transferability.

**Support Vector Machines (SVMs)** is one of the most influential classification algorithms in the 1990s, which was widely used in visual pattern recognition. Denote $(x_1, y_1), (x_2, y_2), \ldots, (x_n, y_n)$ as a sample set where $x_i \in R^m$ is the feature vector while $y_i \in \{-1, 1\}$ is the class label. The intuitive idea of SVMs is to separate the dataset according to their ground truth label, using a hyperplane. Mathematically, $wx + b = 0$. However, for real-world problems with high complexity, a linear classifier is not sufficient to provide an accurate prediction. A kernel function is then introduced to provide non-linearity by transferring the features. The SVMs are first applied in recognizing human action [39], where the features are extracted using local measurement of spatio-temporal interest points.

**Hidden Markov Model (HMM)** assumes that the current state of a random variable depends on its previous state. However, the state cannot be observed directly but can influence the behavior of an observable process. The hidden state space consists of one of $N$ possible values, modeled by a distribution. This implies that the state of the object at time $t$ is selected from one of the hidden states, and a transition matrix is introduced to describe the probability of state at time $t + 1$ given the current state. For each of the $N$ possible states, there is an emission function that illustrates the observed value. During training, the parameters of emission function, as well as the transition matrix, are determined according to the observed features.

## 3.3  Deep Learning Methods

A neural network with a deep structure, which is known as deep model, has a long history and was popular in the 1980s and 1990s [48]. However, due to the limitation of datasets and computation power, its fashion fell out in the 2000s. In recent years, with the emergence of large annotated datasets, such ILSVR [38], PASCAL VOC [9], and the development of high-performance computing techniques, deep models were proved effective by many proposed models, such as VGG16 [42], ResNet [16], etc.

**Object detection** aims to select the objects from a given image. For the large-scale image object detection task, R-CNN [13] introduced an inspiring two-stage architecture by combining a proposal detector and region-wise classifier. SPP-Net [15] and Fast R-CNN [12] were then introduced with the idea of region-wise feature extraction which significantly accelerated the overall detector. Girshick *et al*. [37] proposed a Regional Proposal Network, which is almost cost-free by sharing conv features with a detection network, for object-bound prediction. A multistage detector Cascade R-CNN [5] was then proposed which improved the accuracy of detection by setting increasing IoU thresholds for a sequence of detectors.

**Action recognition** has similar objective with object detection but is performed video-wise. Academic research in action recognition has made great progress in recent years [28]. Karpathy *et al*. [24] studied the performance of CNN and found that CNN architecture is capable of action recognition in large-scale video. Ng *et al*. [47] adopted a CNN for feature extraction, followed by an LSTM for action classification. Simonyan *et al*. [41] proposed a two-stream ConvNet, consisting of separate networks for the frame and optical flow, incorporating spatial and temporal networks. Ji *et al*. [21] proposed CNN-based human detector and head tracker. Tran *et al*. [44] proposed a simple but effective 3D CNN architecture for video classification.

# 4. Tracking Models

The starting point of behavior analysis is tracking the target objects. Prior to the age of deep learning, statistical learning methods dominated the empirical research of Artificial Intelligence.

Smart vivarium [2, 3] is one of the representative works that applies tracking algorithms into the semi-natural enclosure. Different from an open environment, occlusion between objects happens more frequently in an enclosure. To address this issue, the algorithm combines Bayesian Multiple Blob tracker [19] and contour tracker [30]. To increase the robustness of the model, bootstrap filtering is performed using blob observations, and course observations for each mouse independently. At last, the output of the algorithm is computed and sampled independently for each mouse and weighs their importance. The advantage of adopting particle filtering is that the important region for each mouse is calculated independently, preventing loss of track of an occluded object.

One of the most inspiring contributions of the smart vivarium is that it fills the deficiency that prior tracking algorithms have difficulty in detecting objects in occlusion with a lower computational burden. However, it also inherits some of the weaknesses of the original works. For instance, the bounding region does not always match the posture of objects since the templates are fixed. Second, it is sometimes stuck in local optima due to the roughness of the likelihood function of the contour tracker. At last, there is a chance that the identity labels swap when there are multiple occlusions, which is a usual failure mode, that is observed in various algorithms.

The EthoVision [43] a video tracking system that builds on straightforward computer vision methods. The system offers three detection methods: (1) gray scaling separates the image into background and foreground according to the pixel brightness; (2) subtraction method first records a reference image for the background, and then subtracts the value of the reference image from the live image to acquire the object location; (3) the third method defines the color of an object by its hue and saturation and tracking the object by matching the pattern with the given image. However, their method assumes that the background of the video is stationary, and there are limited objects that hardly overlap. BioTracker [35] is an open-source computer vision framework. With a complete video I/O, graphics overlay and mouse and keyboard interfaces, the framework provides a user-friendly environment to those who have limited computer vision knowledge.

Janelia Automatic Animal Behavior Annotator (JAABA) is [23] an intuitive, interactive system for annotating laboratory animals, including mouse, fruit fly and larva. Given a dataset, the interface of JAABA allows users to label a selected animal in a selected frame of behavior. The system then computes per-frame features from the trajectory data as well as windowed features that provide temporal context for the current frame. These features are passed to the JAABA machinery, which learns a new behavior pattern for labeling the future data.

## 4.1 Behavior Detection

Prior to the progress of deep learning methods, research in action detection can be viewed as a two-step process: spatial-temporal feature extraction and action detection based on the extracted features.

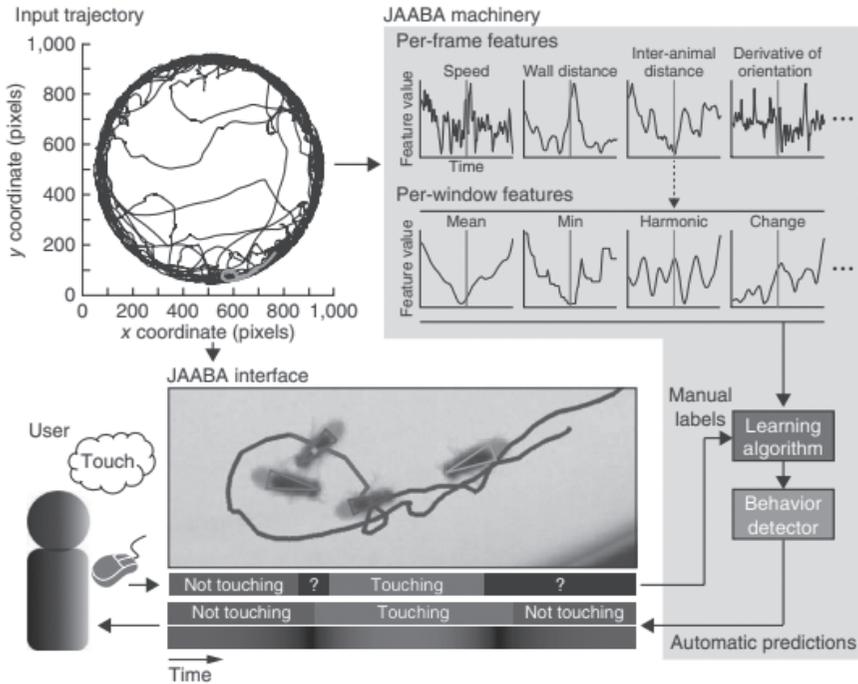

**Fig. 3:** JAABA overview. Top left: is the input trajectory: the *x,y* position of a fly over 1000s of video is plotted in black. *Bottom*: The interface of JAABA where users interact with the interface to annotate the behavior. *Gray shading*: The underlying JAABA machinery where the per-frame features and per-window features are computed and used as inputs for training the detector

Dollar *et al*. [7] is one of the earliest works that illustrate the effectiveness of using spatial-temporal features in mouse behavior detection. They suggest that despite the overall appearance of two instances of the same behavior might be different, the motion of most of the interest points should be similar. Based on that observation, they extract spatial-temporal cuboids that are centered at the interest points and create a collection of cuboids phenotypes using clustering algorithms. Their experiments show that the algorithm has better performance in identifying emotions than detecting mouse behavior, even though the behaviors are contained in the dataset are simple. The following works address the inadequacy of the model by improving the feature extractor, as well as the classification methods.

Hong *et al*. [17] provide a detailed example of applying Artificial Intelligence methods to analyze genetic influences on the social behavior of mice. In particular, they integrate a visual camera with a depth camera to obtain pose information. The information is then used to extract features from body size to relative angles. Although the classier they applied for behavior detection are standard methods with limited modification, they provide a thorough investigation of how genetic differences influence the behavior pattern of the mouse.

Dankert *et al*. [6] propose a method based on machine vision methods to measure the aggression and courtship of fruit flies. Based on the monitored visual data, they introduce a method to compute the location, orientation, and wing posture of each fly. Once the features are computed, they estimate the social behavior using statistical methods. Based on their work, Eyjolfsdottir *et al*. [11] introduced additional features that could better describe the status of interacting flies. Apart from that, they suggested that classifying the actions frame-by-frame can result in a noisy result, and introduced a sliding window that takes the features in adjacent frames into account to increase the robustness of their method. Before the training of the model, the features are then transferred into different features to provide a thorough description of the current status of the objects, including temporal region features, harmonic features, boundary features, etc. The prediction is performed by a hidden semi Markov model. As the deep learning methods shows its potential in various fields, recent attempts have been made to introduce deep neural network into biology research. The advantage is that deep learning models are able to integrate feature extraction and classification into an end-to-end pipeline, providing better transferability. However, it is obtained at the expense of interpretability. Jiang *et al*. [22] are one of the earliest methods that introduce a 2D-3D hybrid framework for detecting social behavior of fruit flies, where the 2D convolutional network is used for extracting the features frame-wise, while the 3D convolutional blocks fuse the spatial-temporal features. However, it is observed that due to the high similarity between different actions, the model performance is not satisfying.

Another example is provided by Eyjolfsdottir *et al*. [10], in which the authors introduce the recurrent neural network to process the sequential motion data. The network has a discriminative part for action classification and a generative part for motion predicting. The recurrent cells of the generative part are laterally connected, which enables the network to learn a high-level representations of behavioral phenomena.

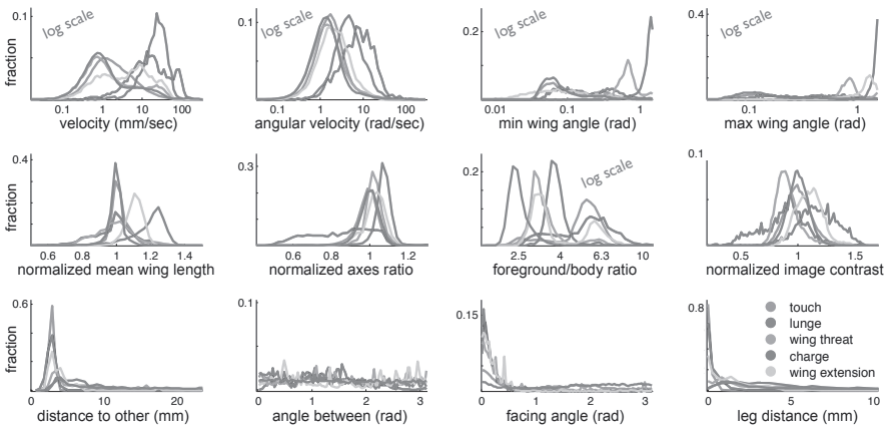

**Fig. 4:** Frame-wise feature distribution for the actions of the boy meets boy sub-dataset from fly-vs-fly [11]. The features are used to train a hidden semi-Markov model

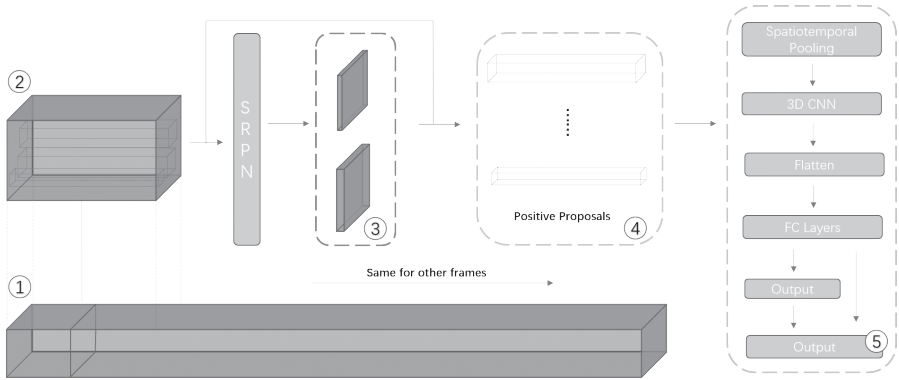

**Fig. 5:** Model structure from [22]. (1) Feature map of a batch. (2) Bout features for classifying behaviors in frame $k + 1$, containing features of frame 1 to $2k + 1$. Likewise, the last bout in the batch is used for classifying behaviors in frame $L - k$, containing features frame $L - 2k$ to frame $L$. Thus, there are $L - 2k$ bouts in a batch of input. (3) Output of SRPN consist of a 'score map' and a 'coordinates map', stands for the probability of 'behavior score' and adjusted bounding boxes coordinates of the corresponding anchor box. (4) Proposed bounding boxes are then sorted according to their 'behavior score', $M_{clx}$ bounding boxes with highest score are selected and then fed into classifier. (5) The classifier starts with spatio-temporal pooling layer to pool features in bounding boxes into a fixed size, followed by 3D convolutional layers and fc layers.

# 5.  Other Applications

## 5.1  Sensor-based Tracking and Detecting

The development of sensor technologies enables a variety of types of data other than visual data. In this section, we provide a discussion on the novel usage of advanced monitor technologies.

Mazur-Milecka and Ruminski [34] propose an efficient deep learning-based method for tracking laboratory mice. The model uses U-net and V-net architectures to convert heat maps to the gray-scale images as well as the segmentation map. However, their model is only valid when the enclosure is clear, otherwise, the heatmap can be affected by other objects.

Weissbrod *et al*. [45] introduce a video radio-frequency-identified (RFID) tagging system and tracking system in a semi-natural environment for tracking laboratory animals and identifying their behavior. RFID technology is able to identify the location and movements of tagged targets and transmit the information to antenna receivers. The collected positional and movement information is then used for social behavior analysis.

However, there are several limitations of such an approach. First, setting up an RFID system requires hardware support, which cannot always be matched for every laboratory. Second, the system is useful in tracking objects, while with only position information, it is hard to provide a straightforward demonstration of detected behaviors.

## 5.2 Pose Estimation

DeepLabCut [32] is one of the first works that applies deep learning methods into the pose estimation. The model is developed from the DeepCut algorithms [36, 18] and shows that a limited number of training data is sufficient to train the model with human-level accuracy.

LiftPose3D [14] is a deep network-based method that reconstructs 3D poses from a single 2D camera view. The model is developed from a network that was initially designed to estimate human posture [31]. Given a 3D poses library, a neural network is trained to match its key points at different 2D angles. Once the matching between key points from the 2D frame and the 3D model is established, the model is able to reconstruct the posture of the target from its 2D image at any angle.

OptiFlex [29] is the first video-based architecture for estimating animal pose. The model consists of a flexible base model that accounts for variability in animal body shape, and an optical flow model that obtains temporal context information from nearby frames. Similar to prior works, the base model is applied to provide a baseline estimation of the target pose, while the novel optical flow module takes a sequence of heatmap frames to enhance the robustness of prediction.